\documentclass[11pt]{article}

\usepackage{amsmath}
\usepackage{amsfonts}
\usepackage{latexsym}
\usepackage{bm}

\usepackage{color}
\usepackage{graphicx}
\usepackage{subcaption}

\usepackage{tikz}
\usetikzlibrary{graphs,arrows,automata}

\newcommand{\be}{\begin{enumerate}}
\newcommand{\ee}{\end{enumerate}}

\usepackage{frcursive}
\usepackage[T1]{fontenc}
\newenvironment{frcseries}{\fontfamily{frc}\selectfont}{}

\def\beq{\begin{equation}}
\def\eeq{\end{equation}}

\def\elabel#1{\label{e:#1}}
\def\eq#1/{(\ref{e:#1})}

\title{A Birth and Death Process for Bayesian Network Structure Inference}

\author {D. Jennings and J. N. Corcoran\\
University of Colorado }

\begin{document}

\maketitle
\vspace{-.8cm}

\begin{abstract} \small \noindent
Bayesian networks (BNs) are graphical models that are useful for representing high-dimensional probability distributions. There has been a great deal of interest in recent years in the NP-hard problem of learning the structure of a BN from observed data. Typically, one assigns a score to various structures and the search becomes an optimization problem that can be approached with either deterministic or stochastic methods. In this paper, we walk through the space of graphs by modeling the appearance and disappearance of edges as a birth and death process and compare our novel approach to the popular Metropolis-Hastings search strategy. We give empirical evidence that the birth and death process has superior mixing properties.

\bigskip

\end{abstract}


\setcounter{page}{0}

\section{Introduction}

Bayesian networks (Pearl \cite{pearl88}) are convenient graphical expressions
for  high dimensional probability distributions representing complex
relationships between a large number of random variables. A Bayesian network
is a \emph{directed acyclic graph} consisting of nodes which represent
random variables and arrows which correspond to probabilistic dependencies
between them.

There has been a great deal of interest in recent years on the NP-hard problem
of learning the structure (placement of directed edges) of Bayesian networks
from data (\cite{chen99},\cite{dobrahans},\cite{friedman03},\cite{friedman00},
\cite{hecktutorial},\cite{dirk},\cite{imoto03},\cite{songbird},\cite{ong02}).
Much of this has been driven by the study of genetic regulatory
networks in molecular biology due to advances in technology and,
specifically, microarray techniques that allow scientists to rapidly
measure expression levels of genes in cells. As an integral part of machine
learning, Bayesian networks have also been used for pattern recognition,
language processing including speech recognition, and credit risk analysis.

Structure learning typically involves defining a network score function and is then, in theory, a straightforward  optimization problem. In practice, however, it is quite a different story as the number of possible networks to be scored and compared grows super-exponentially with the number of nodes. Simple greedy hill climbing with random restarts is understandably inefficient yet surprisingly hard to beat. There have been many deterministic and stochastic alternatives proposed in recent years such as gradient descent, genetic, tempering, and Metropolis-Hastings algorithms. There have been different approaches to the task, including order space sampling and the scoring of graph ``features'' rather than graphs themselves. Several of these methods have offered some improvement  over greedy hill climbing but can be difficult to implement. Deterministic methods tend to get stuck in local maxima and probabilistic methods tend to suffer from slow mixing.

In this paper we consider a new stochastic algorithm in which the appearance and disappearance of edges are modeled as a birth and death process. We compare our algorithm with the popular Metropolis-Hastings algorithm and give empirical evidence that ours has better mixing properties.

\section{Bayesian Networks}
\textit{Bayesian networks} are graphical representations of the relationships between random variables from high dimensional probability distributions. A Bayesian Network on $N$ nodes is a \textit{directed acyclic graph} (DAG) where the nodes (vertices), labeled $1,2,\ldots,N$, correspond to random variables $X_{1}, X_{2}, \ldots, X_{N}$ and directed edges correspond to probabilistic dependencies between them. We say that node $i$ is a {\emph{parent}} of node $j$ and node $j$ is a {\emph{child}} of node $i$ if there exists a directed edge from $i$ to $j$, (in which case we write $i \to j$), and use the notation $pa_{j}$ to denote the collection of all parents of node $j$. We will refer to nodes and their associated random variables interchangeably. Thus, $pa_{j}$ may also represent the collection of parent random variables for $X_{j}$. Rigorously, a Bayesian network consists of a DAG and a set of conditional densities $\{P(X_{i}|pa_{i})\}_{i=1}^{N}$ along with the assumption that the joint density for the $N$ random variables can be written as the product
$$
P(X_{1},X_{2},\ldots,X_{N}) = \prod_{i=1}^{N} P(X_{i} | pa_{i}).
$$
In other words, all nodes are conditionally independent given their parents.

In the problem of structure inference, the DAG is not explicitly known. A set of observations $D = (D_{1},D_{2},\ldots,D_{M})$ is given, where each $D_{i}$ is an $N$-tuple realization of $(X_{1}, X_{2}, \ldots, X_{N})$. The goal is then to recover the ``best'' edge structure of the underlying Bayesian Network, which may be measured in many ways. For example, one may consider the best DAG as the one that maximizes the posterior probability
\beq
\elabel{posterior}
P(G | D) \propto P(D | G) P(G)
\eeq
over $G$. Indeed, (\ref{e:posterior}) is important for many measures of ``best DAGs'' and it is the goal of this paper to simulate DAGs efficiently from this distribution.

Throughout this paper, we will use the common assumption that the data $D$ come from a multinomial distribution, allowing us to analytically integrate out parameters to obtain a score which is proportional to $P(G | D)$.

\section{Jump Processes}
Let $(\Omega,\mathcal{F})$ be a state space consisting of a non-empty set $\Omega$ and a sigma algebra $\mathcal{F}$ on $\Omega$.
A \textit{jump process} on $(\Omega,\mathcal{F})$ is a continuous time stochastic process that is characterized as follows. Assume the process is in some some state $x \in \Omega$.
\begin{itemize}
\item The waiting time until the next jump follows an exponential distribution with rate (or intensity) $\lambda(x)$ and is independent of the past history.
\item The probability that the jump lands the process in $F \in \mathcal{F}$ is given by a transition kernel $K(x,F)$.
\end{itemize}
It is known (e.g. \cite{feller1968introduction}, \cite{preston1975spatial}) that there exists a $Q_t : \Omega \times \mathcal{F} \to \mathbb{R}^+$ so that $Q_t(x,F)$ is the probability that at time $t$ the process is in $F$ given that the process was in state $x$ at time $0$. Such $Q_t$ are defined as the solution to Kolmogorov's backward equation
$$
\frac{\partial}{\partial t} Q_t(x,F) = -\lambda(x) Q_t(x,F) + \lambda(x) \int_\Omega Q_t(y,F) K(x,dy) .
$$

Furthermore, let $Q_t^{(n)}(x,F)$ be the probability of a transition from $x$ to $F$ using at most $n$ jumps. If $\lambda(x)$ is bounded then
$$
Q_{t}^{(\infty)}(x,F) := \lim_{n\to\infty} Q_t^{(n)}(x,F)
$$
is the unique minimal solution to Kolmogorov's forward equation
$$
\frac{\partial}{\partial t} Q_t(x,F) = -\int_F \lambda(z)Q_t(x,dz) + \int_\Omega \lambda(z) K(z,F) Q_t(x,dz).
$$
(It is ``minimal'' in the sense that if $R_{t}(x,F)$ is any other nonnegative solution, then $R_{t}(x,F) \geq Q_{t}^{(\infty)}(x,F)$ for all $t \geq 0$, $x \in \Omega$, and $F \in {\cal{F}}$.)

For a distribution, $\pi$, to be invariant under such a process, $\pi$ must satisfy the detailed balance conditions
$$ \pi(x) \lambda(x) K(x,dy)d\mu(x) = \pi(y)\lambda(y) K(y,dx)d\mu(y) $$
with respect to some density $\mu$.

Preston \cite{preston1975spatial} extended this jump process to the trans-dimensional case where jumps from states in $\Omega_n$ can move the process to a one dimension higher state, living in $\Omega_{n+1}$, with (birth) rate $\lambda_b(x)$ or to a one dimension lower state, living in $\Omega_{n-1}$, with (death) rate $\lambda_d(x)$. Associated with these birth and death rates are birth and death kernels, $K_b$ and $K_d$. The total jump rate and the transition kernel are then given by
$$
\lambda(x) = \lambda_b(x) + \lambda_d(x)
$$
$$
K(x,F) = \frac{\lambda_b(x)}{\lambda(x)} K_b(x,F) + \frac{\lambda_d(x)}{\lambda(x)} K_d(x,F).
$$

For a configuration $x$ with $n$ points to move to a configuration $x^{\prime}$ with $n+1$ points (or vice versa), the detailed balance conditions simplify (\cite{preston1975spatial}, \cite{ripley1977modelling}). In this case one needs the the birth rates to balance the death rates with respect to $\pi$. That is, we require that
$$
\pi(x) b(x,x^{\prime} \setminus x) = \pi(x^{\prime}) d(x^{\prime},x^{\prime} \setminus x)
$$
where $b(x,x^{\prime} \setminus x)$ is the birth rate of the single point $x^{\prime} \setminus x$ given that the current configuration of points is $x$, and $d(x^{\prime},x^{\prime} \setminus x)$ is the death rate of the single point $x^{\prime} \setminus x$ given that the current configuration of points is $x^{\prime}$. These relate to the total birth and death rates in that
$$
\lambda_{b} (x) = \sum b(x, x^{\prime} \setminus x)
$$
$$
\lambda_{d} (x^{\prime}) = \sum d(x^{\prime}, x^{\prime} \setminus x)
$$
where the birth sum is taken over all states $x^{\prime}$ that consist of configuration $x$ with the addition of a single point and the death sum is taken over all states $x$ that consist of configuration $x^{\prime}$ with a single point deleted.

\section{A Birth and Death Process on Edges of a BN}
To construct a jump process for BN structure inference, our goal is to construct a birth and death process acting on edges of a BN which has invariant distribution $P(G | D)$.

The relevant state space is ($\mathcal{G}$,$2^{\mathcal{G}}$), where $\mathcal{G}$ is the set of all DAGs with $N$ nodes and $2^{\mathcal{G}}$ is the power set of ${\mathcal{G}}$. We define the disjoint sets $\mathcal{G}_k$, $k=0,\dots,\frac{N(N-1)}{2}$, to be the set of
DAGs with exactly $k$ edges. Our jump process will then jump between the $\mathcal{G}_k$ for adjacent values of $k$.

For $G \in \mathcal{G}_k$, denote the graph with the addition of the edge from node $i$ to node $j$ by $(G \cup \{i\to j\}) \in \mathcal{G}_{k+1}$, and the graph with the removal of the edge from $i$ to $j$ by $(G \setminus \{i\to j\}) \in \mathcal{G}_{k-1}$. Detailed balance then requires that, for every edge $i\to j$ that is a valid (non-cycle causing) addition,
$$
P(G | D) \, b(G, \{i\to j\}) = P(G \cup \{i\to j\} | D) \, d(G \cup \{i \to j\}, \{ i \to j \}) .
$$
It is convenient to let
$$
d(G \cup \{i \to j\}, \{i \to j\}) = 1
$$
so that
$$
\begin{array}{lcl}
b(G, \{i\to j\}) &=& \frac{P(G \cup \{i\to j\} | D)}{P(G | D)} \\
\\
&=& \frac{P(D | G \cup \{i\to j\}) P(G \cup \{i\to j\})}{P(D | G) P(G)} \\
\end{array}
$$
If we let $\Delta_{j}$ denote the $M$-dimensional vector of observations of $X_{j}$ in the data set $D$, this birth rate may be rewritten as
$$
b(G, \{i\to j\})= \frac{P(\Delta_{j} | \Delta_{pa_j'}, G \cup \{i\to j\}) P(G \cup \{i\to j\})}{P(\Delta_{j} | \Delta_{pa_j}, G) P(G)}.
$$
Here, $\Delta_{pa_{j}}$ is the $M \times k$ matrix of data points for the $k$ parents  of node $j$ in $G$ ($\Delta_{pa_{j}} = \emptyset$ if $k=0$) and $ \Delta_{pa_j'}$ is the $M \times (k+1)$ matrix of data points for the $k$ parents  of node $j$ in $G \cup \{i \rightarrow j\}$.

The transition rates are then given by
$$ \lambda_b(G) = \sum_{\text{valid }i\to j} b(G, \{i\to j\}) ,$$
$$ \lambda_d(G) = \sum_{\{i \to j\} \in G} 1 .$$

With this, we can easily construct a way to simulate from this process in the following way.

\be
\item Start with an arbitrary initial DAG, $G$
\item Compute the birth rates $b(G, \{i \to j\})$ for all possible valid $i \to j$ edge additions to $G$. Compute
$$
\lambda_{b}(G) = \sum_{\text{valid }i\to j} b(G, \{i\to j\})
$$
and
$$
\lambda_d(G) = \sum_{\{i \to j\} \in G} 1.
$$
\item With probability $\lambda_{d}(G)/(\lambda_{b}(G)+\lambda_{d}(G))$, remove a randomly selected existing edge.

Otherwise, add valid edge $i \to j$ with probability $b(G, \{i\to j\})/\lambda_{b}(G)$.

\item Return to step 2.
\ee

At first glance, it may seem like such an algorithm would be computationally expensive, as the required birth rates depend on computing the score for two different graphs. However, if we assume a modular score, the computation at each step is manageable. A modular score means we have
$$ P(D | G) = \prod_{i=1}^N P(\Delta_{i} | \Delta_{pa_i}, G) $$
which leads to a birth rate of
\begin{align*}
b(g, \{i \to j\}) &= \frac{ P(G \cup \{i \to j\}) \prod_{i=1}^N P(\Delta_{i} | \Delta_{pa_i'}, G \cup \{i \to j\})}{ P(G) \prod_{i=1}^N P(\Delta_{i} | \Delta_{pa_i'}, G)} \\
&= \frac{P(G \cup \{i \to j\}) P(\Delta_{j} | \Delta_{pa_j'}, G \cup \{i \to j\})}{P(G) P(\Delta_{j} | \Delta_{pa_j}, G)}
\end{align*}
So, for each birth rate, we only need the ratio of the altered score to current score of a single node corresponding to the child end of the proposed edge.

Further computational relief comes from the fact that most of the birth rates can be stored from previous steps. After adding or removing an edge, the only birth rates that need to be recalculated are for edges pointing to nodes whose parent set has been altered, edges which were not previously valid, or edges which are no longer valid.

These realizations allow for a more complete and efficient algorithm
\be
\item Start with an initial DAG, $G$
\item For all possible edge additions, $i \to j$, calculate the birth rate $b(G,\{i \to j\})$
\item With probability $\lambda_{d}(G)/(\lambda_{b}(G)+\lambda_{d}(G))$, remove a randomly selected existing edge.

Otherwise, add a valid edge $k \rightarrow \ell$ with probability $b(G,\{k \rightarrow \ell\}).$

\item If any of the following are true, update the birth rate $b(G,\{i \to j\})$.
\begin{itemize}
\item $ j = \ell $
\item Edge $i \to j$ was not a valid addition before the addition or removal of edge $k \to \ell$ but now is valid
\item Edge $i \to j$ was a valid addition before the addition or removal of edge $k \to \ell$ but now is not longer valid
\end{itemize}
\item Return to step 3.
\ee

\section{Experimental Results}

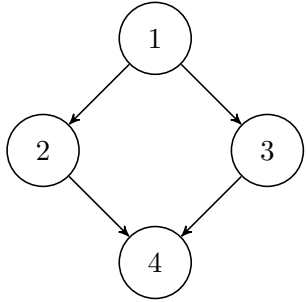
\begin{figure}
\begin{center}
\begin{tikzpicture}[->,>=stealth',auto,node distance=60pt,semithick]

\node[state] (A) {$1$};
\node[state] (B) [below left of=A] {$2$};
\node[state] (C) [below right of=A] {$3$};
\node[state] (D) [below left of=C] {$4$};

\path (A) edge (B);
\path (A) edge (C);
\path (B) edge (D);
\path (C) edge (D);

\end{tikzpicture}
\end{center}
\caption{4 Node Graph Used for Testing}
\label{fig:fournode}
\end{figure}

While the theory guarantees that our edge birth and death process will have the correct stationary distribution, in this Section we investigate the mixing time and whether or not Monte Carlo simulation of graphs from $P(G|D)$ using our method is feasible in practice. We first consider a simple 4 node graph so that we may compare our results with exact computations. We generated 50 observations of $X_{1}, X_{2}, X_{3}, X_{4}$ as related by the DAG in Figure \ref{fig:fournode}. Each node was allowed to take on values in $\{1,2,3,4\}$, with equal probability.

\begin{figure}
\begin{subfigure}{0.5\textwidth}
\includegraphics[width=\linewidth]{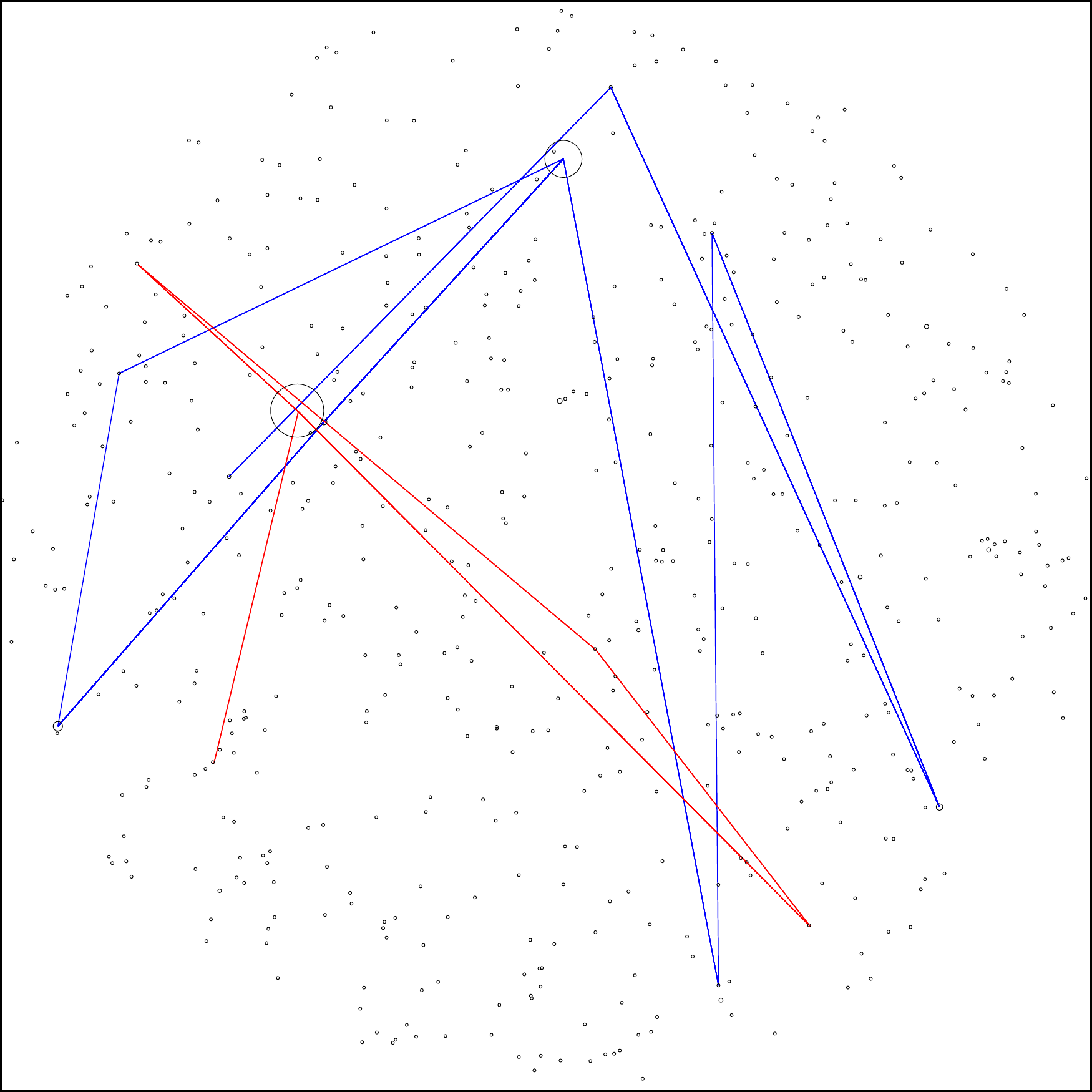}
\label{fig:pathsmh}
\end{subfigure}
\begin{subfigure}{0.5\textwidth}
\includegraphics[width=\linewidth]{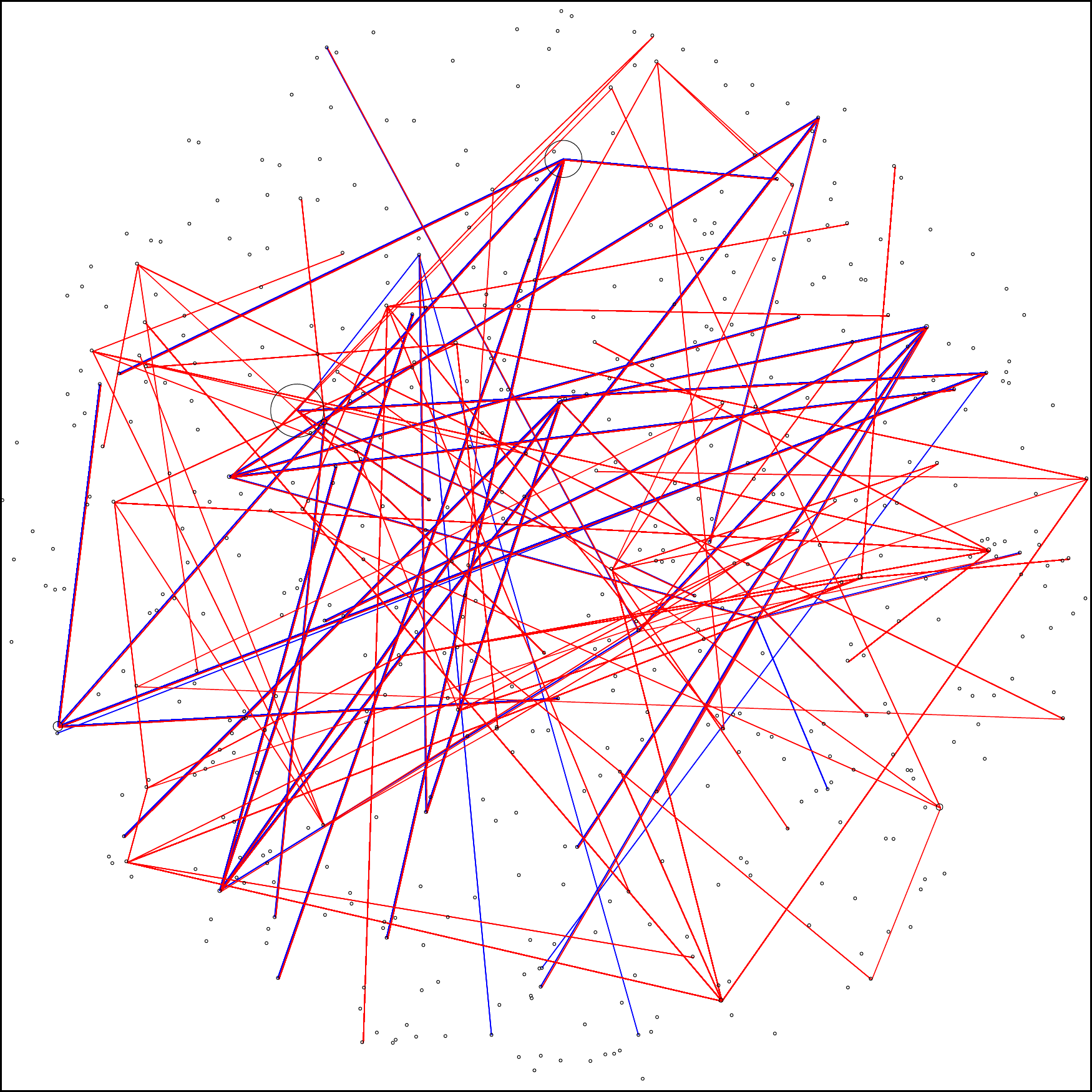}
\label{fig:pathsbd}
\end{subfigure}
\caption{(a) Paths of two (red and blue) independent Metropolis Hastings chains of 20000 steps. (b) Paths of two independent edge birth and death process runs of 1500 jumps. The circles represent each of the 543 possible DAGs, with size proportional to the posterior probability of the graph.}
\label{fig:paths}
\end{figure}

We then ran both the standard Metropolis Hastings chain for BNs (\cite{madiganyork1995}) and our edge birth and death algorithm. Figure \ref{fig:paths} shows two independent runs of each algorithm. While both algorithms tend to find high probability regions of the state space, our edge birth and death algorithm explores much more of the state space.

One benefit of running Monte Carlo methods is they allow us to easily calculate the probabilities of specific graph features, e.g. edge probabilities. With the edge birth and death process, the edge probabilities correspond to the proportion of time that the graphs contain the given edge. With the same 4 node graph as before, we calculated the edge probabilities and compare them to the true edge probabilities both for $m=100$ observations and $m=500$ observations. The results are shown in Table \ref{tab:edgeprob}.

\begin{table}

\begin{subtable}{0.5\textwidth}
\centering
\begin{tabular}{|c|c|c|c|c|}
\hline
Node & 1 & 2 & 3 & 4 \\
\hline
1 & -- & .06 & .02 & .05 \\
\hline
2 & .06 & -- & .00 & .01 \\
\hline
3 & .02 & .04 & -- & .01 \\
\hline
4 & .04 & .04 & .00 & -- \\
\hline
\end{tabular}
\label{tab:ep100}
\end{subtable}
\begin{subtable}{0.5\textwidth}
\centering
\begin{tabular}{|c|c|c|c|c|}
\hline
Node & 1 & 2 & 3 & 4 \\
\hline
1 & -- & .03 & .02 & .00 \\
\hline
2 & .03 & -- & .00 & .00 \\
\hline
3 & .02 & .00 & -- & .00 \\
\hline
4 & .00 & .00 & .00 & -- \\
\hline
\end{tabular}
\label{tab:ep500}
\end{subtable}
\caption{Magnitudes of errors between estimated and exact probabilities of edges in a 4 node Bayesian network from one run of the edge birth and death algorithm for $100$ observations (left) and $500$ observations (right).}
\label{tab:edgeprob}
\end{table}

\begin{figure}
\begin{subfigure}{0.5\textwidth}
\includegraphics[width=\linewidth]{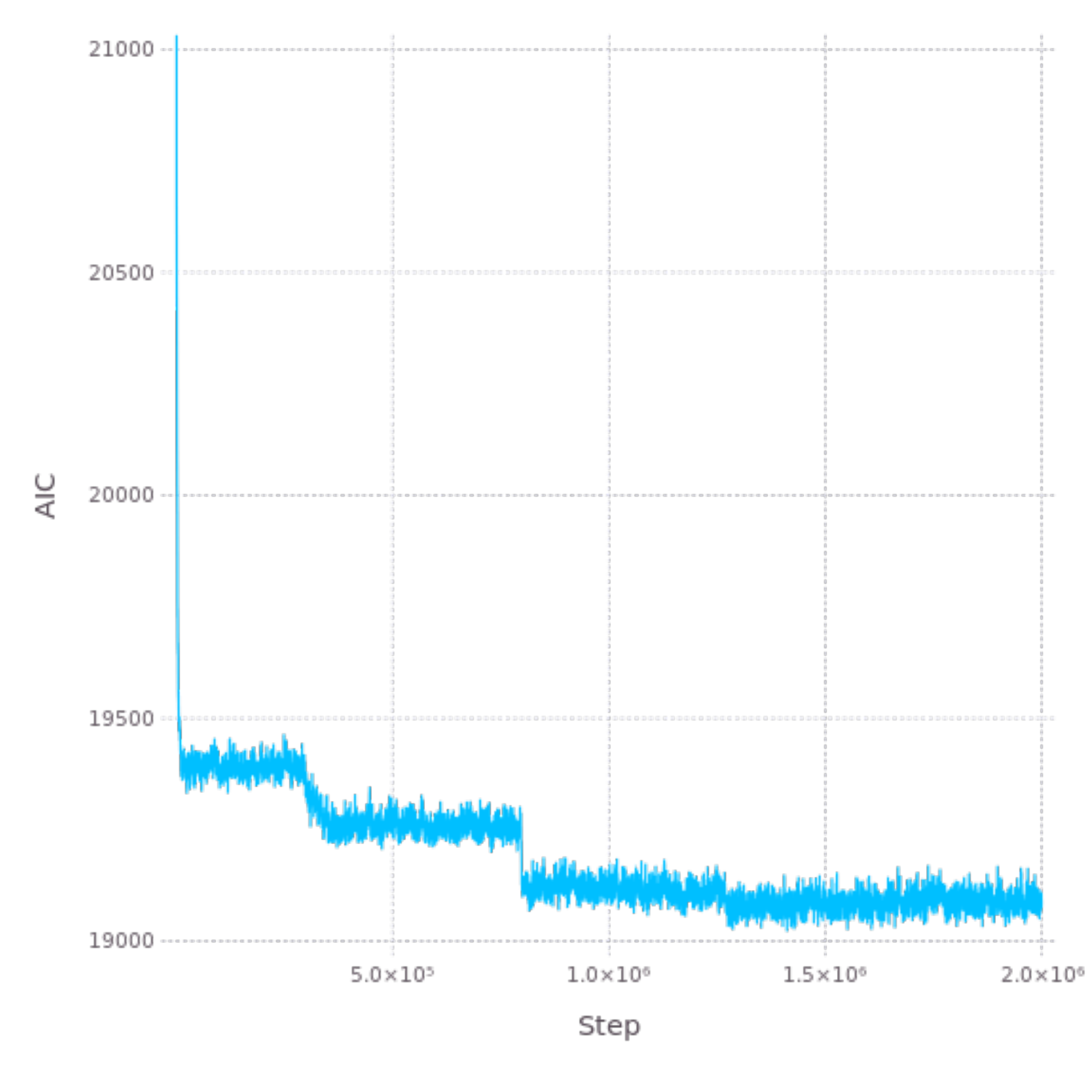}
\label{fig:alarmaicmh}
\end{subfigure}
\begin{subfigure}{0.5\textwidth}
\includegraphics[width=\linewidth]{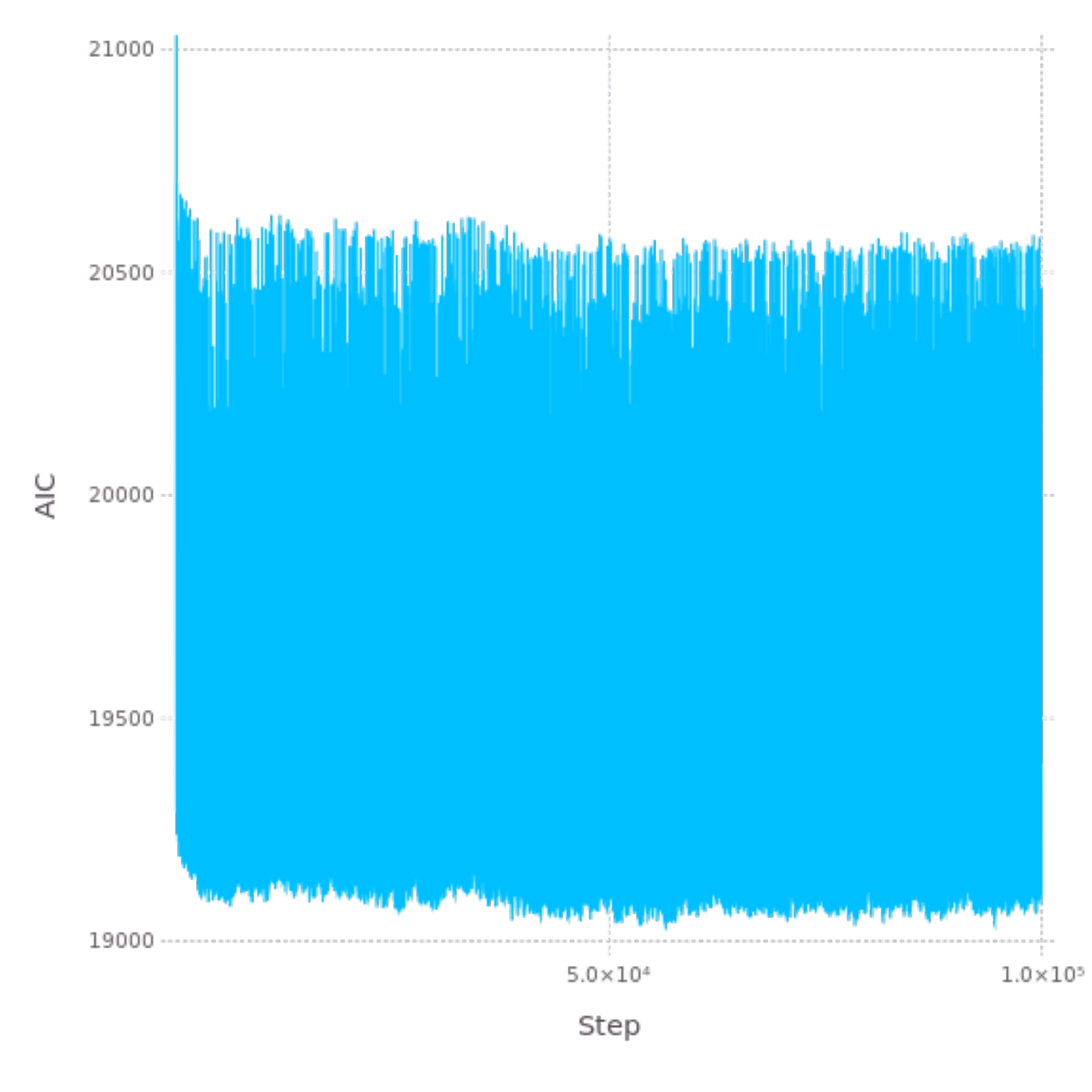}
\label{fig:alarmaicbd}
\end{subfigure}
\caption{(a) AIC score of a Metropolis Hastings run of $2\times 10^6$ steps. (b) AIC score of an edge birth and death process run of $10^5$ steps.}
\label{fig:alarmaic}
\end{figure}

Next, we tested our edge birth and death algorithm on the ``Alarm data set'' compiled by Herskovits (\cite{herskovits}). This data set, often used as a benchmark for structure learning algorithms, consists of 1000 observations of 37 random variables, each taking on 4 possible values. We ran our edge birth and death algorithm for $10^5$ time steps. For comparison, we ran the standard Metropolis-Hastings scheme for $2 \times 10^6$ steps which took approximately the same amount of CPU time. Akaike's information criteria (AIC) was recorded at each step and is plotted for one of these runs in Figure \ref{fig:alarmaic}. As is well known, the MH scheme is prone to getting trapped in local minima, and takes many steps to escape these points whereas our edge birth and death algorithm appears to mix more easily.

\section{Conclusions}
We have presented a new algorithm for sampling from the posterior distribution of Bayesian Networks given data based on a birth and death process. This new edge birth and death algorithm allows for probabilistic inferences to be made about Bayesian Network structures while avoiding some of the downfalls of existing methods. In particular, our edge birth and death process does not get trapped in local extrema as easily as structure MCMC and allows for less restrictive choices of priors over graphs compared to order MCMC.

An open question related to this new algorithm is whether it can be made perfect by applying similar constructions to perfect algorithms for spatial point processes.

\bibliographystyle{plain}

\bibliography{/home/corcoran/BIBS/strings,/home/corcoran/BIBS/masterbib}

\end{document}